\documentclass[10pt,twocolumn,letterpaper]{article}

\usepackage{iccv}
\usepackage{times}
\usepackage{epsfig}
\usepackage{graphicx}
\usepackage{amsmath}
\usepackage{amssymb}
\usepackage{multicol}
\usepackage{multirow}
\usepackage{arydshln}
\usepackage{pifont}

\usepackage{booktabs}
\usepackage{float}
\usepackage[pagebackref=true,breaklinks=true,letterpaper=true,colorlinks,bookmarks=false]{hyperref}

\usepackage[accsupp]{axessibility}  % Improves PDF readability for those with disabilities.

\newcommand{\ignore}[1]{}

\iccvfinalcopy % *** Uncomment this line for the final submission

\def\iccvPaperID{7048} % *** Enter the ICCV Paper ID here

\begin{document}

%%%%%%%%% TITLE
\title{ISD: Self-Supervised Learning by Iterative Similarity Distillation}

\author{\fontsize{11}{11} \selectfont Ajinkya Tejankar$^{1,}\footnotemark[1]\quad$ 
 Soroush Abbasi Koohpayegani$^{1,}\thanks{Equal contribution}\quad$ Vipin Pillai$^{1}\quad$  Paolo Favaro$^{2}\quad$ Hamed Pirsiavash$^{1,3}\quad$   \\
 \\
\fontsize{11}{11} \selectfont $^{1}$University of Maryland, Baltimore County$\quad$ $^{2}$University of Bern$\quad$ $^{3}$University of California, Davis\\
%{\tt\fontsize{9}{9} \selectfont \{soroush,at6,vp7,hpirsiav\}@umbc.edu \quad favaro@inf.unibe.ch}


}

\maketitle

%%%%%%%%% ABSTRACT
\begin{abstract}
Recently, contrastive learning has achieved great results in self-supervised learning, where the main idea is to pull two augmentations of an image (positive pairs) closer compared to other random images (negative pairs). We argue that not all negative images are equally negative. Hence, we introduce a self-supervised learning algorithm where we use a soft similarity for the negative images rather than a binary distinction between positive and negative pairs. We iteratively distill a slowly evolving teacher model to the student model by capturing the similarity of a query image to some random images and transferring that knowledge to the student. Specifically, our method should handle unbalanced and unlabeled data better than existing contrastive learning methods, because the randomly chosen negative set might include many samples that are semantically similar to the query image. In this case, our method labels them as highly similar while standard contrastive methods label them as negatives. Our method achieves comparable results to the state-of-the-art models. Our code is available here: \textcolor{magenta}{\href{https://github.com/UMBCvision/ISD}{https://github.com/UMBCvision/ISD}}.
\end{abstract}

\begin{figure*}
\begin{center}
\end{center}
   \includegraphics[width=1\linewidth]{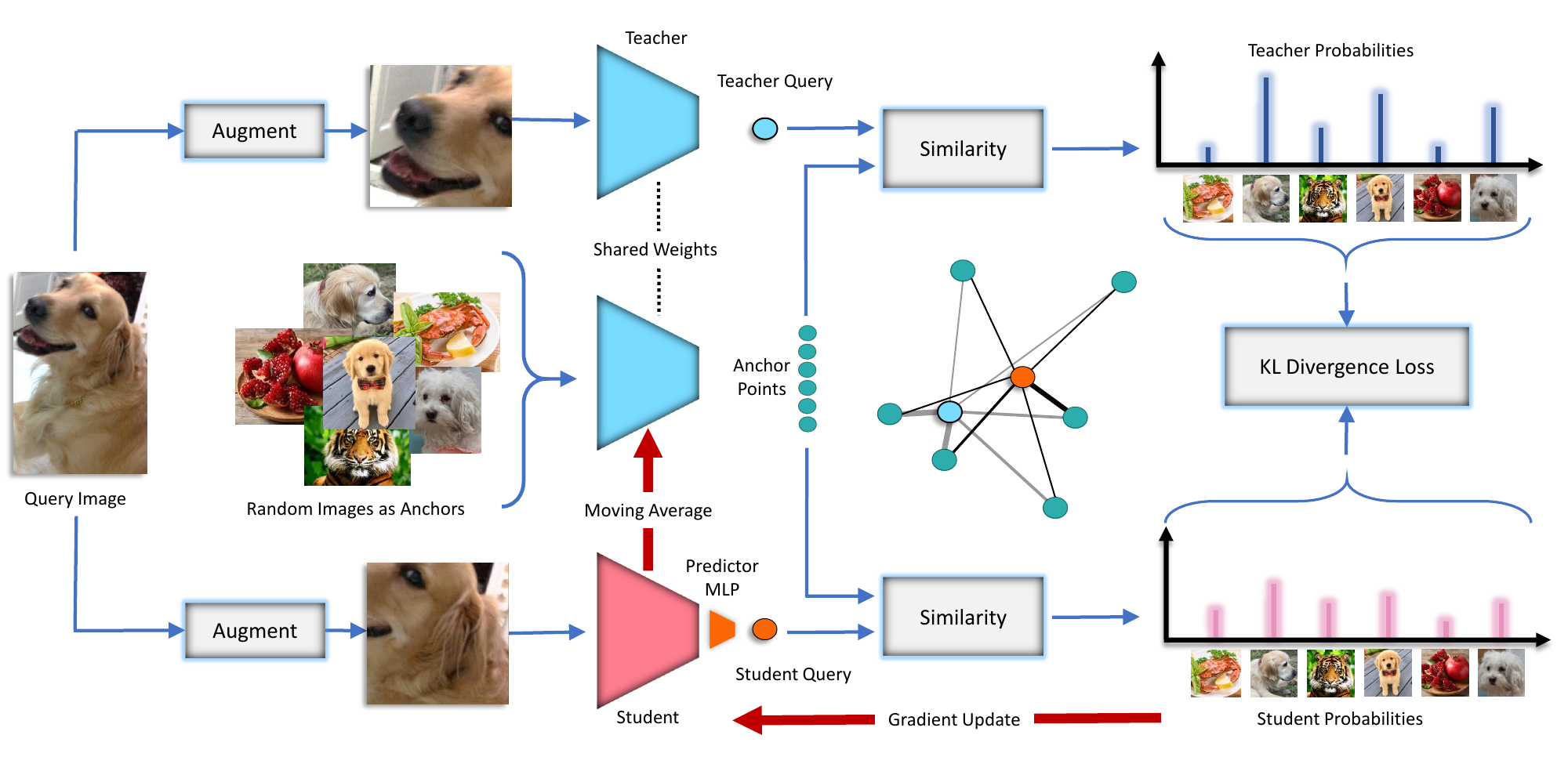}
   \caption{\textbf{Our method:} We initialize both teacher and student networks from scratch and update the teacher as running average of the student. We feed some random images to the teacher, and feed two different augmentations of a query image to both teacher and student. We capture the similarity of the query to the anchor points in the teacher's embedding space and transfer that knowledge to the student. We update the student based on KL divergence loss and update the teacher to be a slow moving average of the student. This can be seen as a soft version of MoCo \cite{he2020momentum} which can handle negative images that are similar to the query image. Note that unlike contrastive learning and BYOL \cite{grill2020bootstrap}, we never compare two augmentations of the query images directly (positive pair).}
\label{fig:teaser}
\end{figure*}

%%%%%%%%% BODY TEXT
\section{Introduction}

We can view the recent crop of SSL methods as iterative self-distillation where there is a teacher and a student. Both teacher and student improve simultaneously while the teacher is evolving more slowly (running average) compared to the student: (1) In the case of contrastive methods e.g. MoCo \cite{he2020momentum}, we classify images to positive and negative pairs in the binary form. (2) In the case of clustering methods (DC-v2 \cite{caron2020unsupervised}, SwAV \cite{caron2020unsupervised}, SeLA \cite{asano2020self}), the student predicts the quantized representations from the teacher. (3) In the case of BYOL \cite{grill2020bootstrap}, the student simply regresses the teacher's embeddings vector. Here, we introduce a novel method using similarity based distillation to transfer the knowledge from the teacher to the student. We argue that our method is more regularized compared to prior work and improves the quality of the features in transfer learning.

In the standard contrastive setting, \eg, MoCo \cite{he2020momentum}, there is a binary distinction between positive and negative pairs, but in practice, many negative pairs may be from the same category as the positive one. Thus, forcing the model to classify them as negative is misleading. This can be more important when the unlabeled training data is unbalanced, for example, when a large portion of images are from a small number of categories. Such scenario can happen in applications like self-driving cars, where most of the data is just repetitive data captured from a high-way scene with a couple of cars in it. In such cases, the standard contrastive learning methods will try to learn features to distinguish two instances of the most frequent category that are in a negative pair, which may not be helpful for the down-stream task of understanding rare cases.
    
We are interested in relaxing the binary classification of contrastive learning with soft labeling, where the teacher network calculates the similarity of the query image with respect to a set of anchor points in the memory bank, converts that into a probability distribution over neighboring examples, and then transfers that knowledge to the student, so that the student also mimics the same neighborhood similarity. In the experiments, we show that our method is competitive with SOTA self-supervised methods on ImageNet and show an improved accuracy when trained on unbalanced, unlabeled data (for which we use a subset of ImageNet).
    
Our method is different from BYOL \cite{grill2020bootstrap} in that we are comparing the query image with other random images rather than only with a different augmentation of the same query image. We believe our method can be seen as a more relaxed version of BYOL. Instead of imposing that the embedding of the query image should not change at all due to an augmentation (as done in BYOL), we are allowing the embedding to vary as long as its neighborhood similarity does not change. In other words, the augmentation should not change the similarity of the image compared to its neighboring images. This relaxation lets self-supervised learning focus on what matters most in learning rich features rather than forcing an unnecessary constraint of no change at all, which is difficult to achieve. 
    
Our distillation method is inspired by the CompRess method \cite{abbasi2020compress}, which introduces an analogous similarity-based distillation method to compress a deeper self-supervised model to a smaller one and get better results compared to training the small model from scratch. Our method is different from \cite{abbasi2020compress} in that in our case, both teacher and student share the same architecture, we do self-supervised learning from scratch rather than compressing from another deeper model, and also the teacher evolves over time as a running average of the student rather than being frozen as in \cite{abbasi2020compress}. %\ajinkya{add seed here as well?}

\ignore{
    %(called ``target'' in BYOL and ``key encoder'' in MoCo)
\begin{itemize}
    \item SimMatch does soft contrast unlike clustering and contrastive methods while also looks at other data points unlike BYOL. Thus, SimMatch is the sweet spot method. The temperature paramter can let it behave like clustering methods and sim matching like BYOL.
    \item It is important to compare the methods in non hw intensive and comparable settings. Under these conditions, is BYOL really better than MoCo?
    \item SwAV shows that the effect of augmentation is orthogonal and the accuracy can always be boosted with improved augmentation.
    \item Self-distillation can also be thought of as way to remove augmentation bias. By avoiding heavy augmentation in the second phase and initializing the student from scratch.
    \item BYOL cannot make use of memory bank which is available even on cheap and non-specialized hardware. Our method can make use of it and be better than BYOL in small batch setting. For large batch, we know that BYOL performs well, but how does it compare to our method? We don't know. Large batch might benefit our method as well. That is a different discussion. Our method is definitely more accessible and easily scalable.
\end{itemize}
}
%------------------------------------------------------------------------

\begin{figure*}
\begin{center}
\end{center}
   \includegraphics[width=1.0\linewidth]{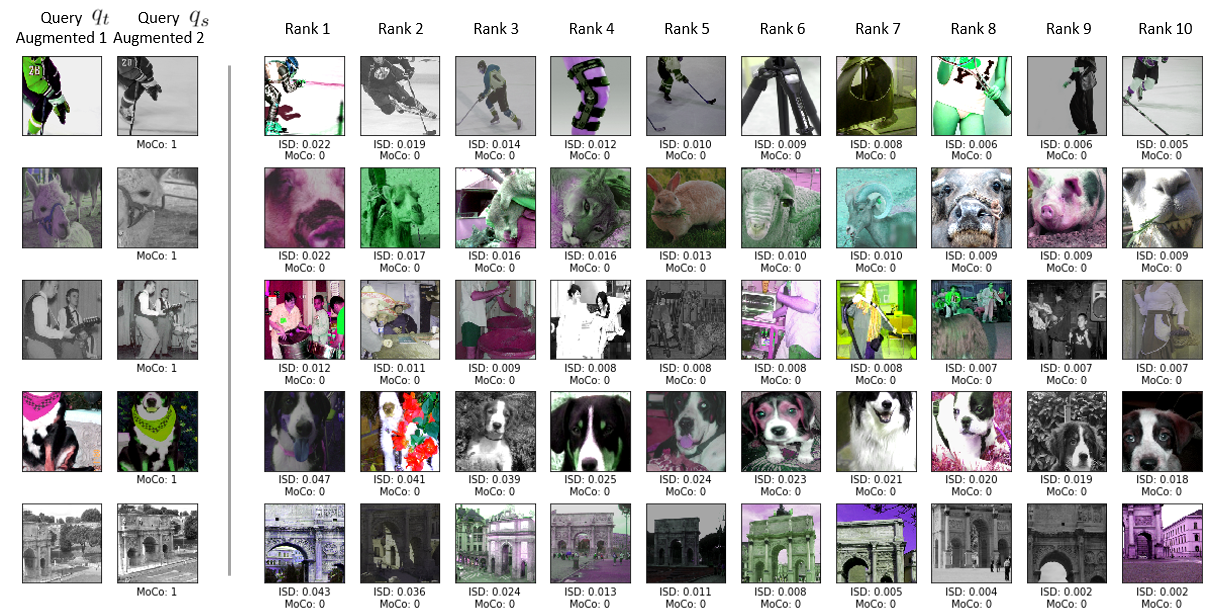}
   \caption{\textbf{Positives vs. negatives:} We sample some query images randomly (left column), calculate their teacher probability distribution over all anchor points in the memory bank (size=128K) and rank them in descending order (right columns). The second left column is another augmented version of the query image that contrastive learning methods use for the positive pair. Our students learns to mimic the probability number written below each anchor image while contrastive learning method (e.g., MoCo) learn to predict the one-hot encoding written below the images. Note that there are lots of images in the top anchor points that are semantically similar to the query point that MoCo tries to discriminate them from the query while our method does not.}

\label{fig:moco_vs_isd}
\end{figure*}

\section{Method}
We are interested in learning rich representations from unlabeled data. We have a teacher network and a student network. We initialize both models from scratch and update the teacher to be a slower version of the student: we use the momentum idea from MoCo in updating the teacher so that it is running average of the student. The method is described in Figure \ref{fig:teaser}. Following the notation in \cite{abbasi2020compress}, at each iteration, we pick a random query image and a bunch of random other images that we call anchor points. We augment those images and feed them to the teacher model to get their embeddings. Then, we augment the query again independent of the earlier augmentation and feed it to the student model only. We calculate the similarity of the query point compared to the anchor points in the teacher's embedding space and then optimize the student to mimic the same similarity for the anchor point at the student's embedding space. Finally, we update the teacher with a momentum to be the running average of the student similar to MoCo and BYOL. Note that our method is closely related to ComPress method \cite{abbasi2020compress} which uses the similarity distillation for compressing a frozen larger model to a smaller one.

More formally, we assume a teacher model $t$ and a student model $s$. Given a query image $q$, we augment it twice independently to get $q_t$ and $q_s$. We also assume a set of $n$ augmented random images $\{x_i\}_{i=1}^n$. We feed $\{x_i\}_{i=1}^n$ to the teacher model to get their embeddings $\{t(x_i)\}_{i=1}^n$ and call them anchor points. We also feed $q_t$ to the teacher and $q_s$ to the student to get $t(q_t)$ and $s(q_s)$ respectively. Then, we calculate the similarity of the query embedding $t(q_t)$ compared to all anchor points, divide by a temperature, and convert to a probability distribution using a SoftMax operator to get:

\vspace{-.1in}
$$p_t(i) = -\text{log} \frac{\text{exp}\big(\text{sim}\big(t(q_t), t(x_i)\big)/\tau_t\big)}{\sum_{j=1}^n \text{exp}\big(\text{sim}\big(t(q_t), t(x_j)\big)/\tau_t\big)}$$

\noindent where $\tau_t$ is the teacher's temperature parameter and $\text{sim}(.,.)$ refers to the similarity between two vectors. In our experiments, we use cosine similarity which is standard in most recent contrastive learning methods.

Then, we calculate a similar probability distribution for the student's query embedding to get:

$$p_s(i) = -\text{log} \frac{\text{exp}\big(\text{sim}\big(s(q_s), t(x_i)\big)/\tau_s\big)}{\sum_{j=1}^n \text{exp}\big(\text{sim}\big(s(q_s), t(x_j)\big)/\tau_s\big)}$$

Where, $\tau_s$ is the student's temperature parameter. Finally, we optimize the student only by minimizing the following loss: 
$$L = KL(p_t || p_s)$$

\noindent and the teacher is updated using the following rule:

$$ \theta_t \leftarrow m \theta_t + (1-m) \theta_s$$

\noindent where $\theta$ refers to the parameters of a model and $m$ is a the momentum hyperparameter that is set be close to one ($0.99$ in our ResNet50 experiments) as in MoCo. Since the teacher is not optimized by the loss directly, the loss can be simplified as cross entropy loss instead of KL divergence. 

Note that unlike the positive pair in contrastive learning methods and BYOL, the query image is never compared to its own augmentation as it is not included in the anchor points. We do this since when the features are mature, the similarity of the query to itself will be very large and will dominate the whole probability distribution. Note that one can convert our method to MoCo by including the query in anchor points and replacing the probability of the teacher $p_t$ with a one-hot encoding vector in which the positive pair (query) corresponds to one and all other anchor points correspond to zero. Figure \ref{fig:moco_vs_isd} shows some example teacher probabilities for both ISD and MoCo.

Our method can benefit from a large number of anchor points to cover the neighborhood of any query image, and also the anchor points are fed to the teacher only that evolves slowly. Hence, we use MoCo's trick of a large memory bank (queue) for the anchor points. The queue size is $128$K in our experiments which uses only 1.6\% of the total memory and less than 1\% of the total computation.

{\bf Different Temperature for student and teacher:} 
Since the student is learning from the teacher, we can use a lower temperature for the teacher compared to the student to make the teacher more confident. In the extreme case, when the teacher uses zero temperature, its output will be a one-hot encoding over the anchor points which is a very sharp distribution. In the experiments we observe best results when the teacher has 10 times smaller temperature.

%We tried a version of ISD where teacher has smaller temperature compare to student. it helps reduce student entropy and force student to do a form of clustering.

\begin{table*}[t!]
\begin{center}
\scalebox{0.90}{
\begin{tabular}{|l|c|c|c|c|c|c|c|}
\hline 
Method & Ref & Batch & Epochs & Sym. Loss & Top-1 & NN & 20-NN \\

& & Size & & 2x FLOPS & Linear & & \\\hline 
\multicolumn{8}{|c|}{ResNet-50} \\
\hline
Supervised & - & 256 & 100 & - & 76.2 & 71.4 & 74.8 \\

\hdashline
% SimCLR\cite{chen2020simple} & \cite{chen2020exploring} & 4096 & 200 & \ding{55} & 68.3 & - & -\\
SwAV \cite{caron2020unsupervised} & \cite{chen2020exploring} & 4096 & 200 & \ding{51} & 69.1 & - & - \\
SimCLR\cite{chen2020simple} & \cite{chen2020simple} & 4096 & 1000 & \ding{51} & 69.3 & - & - \\
MoCo-V2 \cite{he2020momentum} & \cite{chen2020exploring} &  256 & 200  & \ding{51} & 69.9 & - & - \\
SimSiam \cite{chen2020exploring} & \cite{chen2020exploring} & 256 & 200 & \ding{51} & 70.0 & - & - \\
BYOL \cite{grill2020bootstrap} & \cite{chen2020exploring} & 4096 & 200 & \ding{51} & 70.6 & - & -\\
MoCo-V2 \cite{chen2020mocov2} & \cite{chen2020mocov2} & 256 & 400 & \ding{51} & 71.0 & - & - \\
MoCo-V2 \cite{he2020momentum} & \cite{he2020momentum} & 256 & 800 & \ding{55} & 71.1 & 57.3 & 61.0 \\
CompRess* \cite{abbasi2020compress} & \cite{abbasi2020compress} & 256 & 1K+130 & \ding{55} & 71.9 & 63.3& 66.8\\
BYOL \cite{grill2020bootstrap} & \cite{grill2020bootstrap} & 4096 & 1000 & \ding{51} & 74.3 & 62.8 & 66.9 \\
SwAV \textsuperscript{$\dagger$} \cite{caron2020unsupervised} & \cite{caron2020unsupervised} & 4096 & 800 & \ding{51} & \textbf{75.3} & - & - \\
\hdashline
MoCo-V2 \cite{he2020momentum} & \cite{chen2020exploring} &  256 & 200  & \ding{55} & 67.5 & - & -\\
CO2 \cite{wei2020co2} & \cite{wei2020co2} & 256 & 200 & \ding{55} & 68.0 & - & - \\
BYOL-asym & - & 256 & 200 & \ding{55} & 69.3 & 55.0 & 59.2\\
MSF \textsuperscript{$\ddagger$} \cite{koohpayegani2021mean}  & \cite{koohpayegani2021mean} & 256 & 200 & \ding{55} & \bf{72.4} & \textbf{62.0} & \textbf{64.9} \\
ISD & - & 256 & 200 & \ding{55} & 69.8 & 59.2 & 62.0 \\
\hline
\multicolumn{8}{|c|}{ResNet-18} \\\hline
Supervised & - & 256 & 100 & - & 69.8 & 63.0 & 67.6 \\ \hdashline
MoCo-V2 \cite{he2020momentum} & \cite{chen2020exploring} &  256 & 200  & \ding{55} & 51.0 & 37.7 & 42.1 \\
BYOL-asym & - & 256 & 200 & \ding{55} & 52.6 & 40.0 & 44.8\\

ISD & - & 256 & 200 & \ding{55} & \bf{53.8} & \bf{41.5} & \bf{46.6} \\

\hline
\end{tabular}
}

\end{center}

\caption{\textbf{Evaluation on full ImageNet: } We compare our method with other state-of-the-art SSL methods by evaluating the learned features on the full ImageNet. A single linear layer is trained on top of a frozen backbone. Note that methods using symmetric losses use $2\times$ computation per mini-batch. Thus, it is not fair to compare them with the asymmetric loss methods. Further, we find that given a similar computational budget, both asymmetric MoCo-V2 (400 epochs) and symmetric MoCo-V2 (800 epochs) have similar accuracies ($71.0$ vs $71.1$). Under similar resource constraints, our method performs competitively with other state-of-the-art methods. * is compressed from ResNet-50x4. $\dagger$: SwAV is not comparable as it uses multiple crops together. $\ddagger$: is our concurrent (future of this!) work. %Thanks for CVPR rejecting reviewers.
}

\label{tab:imagenet}

\end{table*}

\section{Experiments}

We describe various experiments and their results in this section. We compare our proposed self-supervised method with other state-of-the-art methods on ImageNet and transfer learning. We also demonstrate the advantage of our method compared to MoCo on unbalanced, unlabeled dataset.
% Next, we show how the method can be modified to remove unwanted augmentation biases.

\textbf{Implementation details: } For all experiments, we use PyTorch with SGD optimizer (momentum = $0.9$, weight decay = $1e-4$, batch size = 256) except when stated otherwise. Details about the specific architecture, epochs for training, and learning rate are described for each experiment in its respective section. We follow the evaluation protocols in \cite{abbasi2020compress} for nearest neighbor (NN) and linear layer (Linear) evaluation. We use the ImageNet labels only in the setting of evaluating the learned features. To evaluate how SSL features transfer to new tasks, we perform Linear layer evaluation on different datasets including Food101 \cite{food101}, SUN397 \cite{sun397}, CIFAR10 \cite{cifar}, CIFAR100 \cite{cifar}, Cars \cite{carsdataset}, Flowers \cite{flowers}, Pets \cite{pets}, Caltech-101 \cite{caltech101} and DTD \cite{dtd}. More details about the datasets and training can be found in the  appendix. We follow \cite{grill2020bootstrap} setting for transfer learning and reproduce BYOL results for fairness.

\subsection{Self-supervised learning}
\textbf{BYOL-asym (baseline).} ResNet-50 is recently used as a benchmark in the community. Unfortunately, we cannot run it for 1000 epochs because of resource constraints. Some methods \cite{grill2020bootstrap,caron2020unsupervised} are even slower as they forward the mini-batch through the model more than once. For instance, BYOL method forwards the images twice to calculate the symmetric loss, so $100$ epochs of symmetric BYOL is equivalent to almost $200$ epochs of asymmetric BYOL in terms of running time. As shown in \cite{chen2020exploring}, given a constant budget, there is no big difference between symmetric and asymmetric losses. 
Thus, for a fair comparison with our method and MoCo, we use asymmetric loss, a small batch size (256), momentum for the teacher is $0.99$, and train for 200 epochs. We implement BYOL in PyTorch following \cite{grill2020bootstrap}. We call this baseline as BYOL-asym since it's asymmetric version of BYOL. For ResNet18, hidden units in the MLP for projection and prediction layers is $1024$, and output embedding dimension is $128$. For ResNet50, hidden units in the MLP for projection and prediction layers is $4096$, and output embedding dimension is $512$. We use cos learning rate scheduler with initial learning rate of $0.05$.

\begin{table*}[t!]
    \begin{center}
    \scalebox{0.95}{
    \begin{tabular}{|l|c|c|c|c|c|c|c|c|c|c|c|c|c|}
    \hline
    Method & Ref. & Epochs & Food & CIFAR & CIFAR & SUN & Cars & DTD & Pets & Caltech & Flowers & Mean \\
    & & & 101 & 10 & 100 & 397 & 196 &  &  & 101 & 102 & \\
    \hline
    \multicolumn{14}{|c|}{ResNet-50} \\\hline
    Sup-IN & \cite{grill2020bootstrap} & & 72.3 & 93.6 & 78.3 & 61.9 & 66.7 & 74.9 & 91.5 & 94.5 & 94.7 & 80.9 \\
    \hline
    SimCLR \cite{chen2020simple} & \cite{grill2020bootstrap} & 1000 & 72.8 & 90.5 & 74.4 & 60.6 & 49.3 & 75.7 & 84.6 & 89.3 & 92.6 & 68.6 \\
    MoCo v2 \cite{chen2020mocov2} & - & 800 & 72.5 & 92.2 & 74.6 & 59.6 & 50.5 & 74.4 & 84.6 & 90.0 & 90.5 & 76.5 \\
    BYOL \cite{grill2020bootstrap} & rep. & 1000 & \textbf{75.4} & \textbf{92.7} & 78.1 & 62.1 & 67.1 & \textbf{76.8} & 89.8 & 92.2 & 95.5 & 81.1 \\
    BYOL \cite{grill2020bootstrap} & \cite{grill2020bootstrap} & 1000  & 75.3 & 91.3 & \textbf{78.4} & \textbf{62.2} & \textbf{67.8} & 75.5 & \textbf{90.4} & \textbf{94.2} & \textbf{96.1} & \textbf{81.2} \\
    \hline
    BYOL-asym \cite{grill2020bootstrap} & - & 200 & 70.2 & 91.5 & 74.2 & 59.0 & 54.0 & 73.4 & 86.2 & 90.4 & \textbf{92.1} & 76.8 \\
    MoCo v2 \cite{chen2020mocov2} & - & 200 & 70.4 & 91.0 & 73.5 & 57.5 & 47.7 & \textbf{73.9} & 81.3 & 88.7 & 91.1 & 75.0 \\
    MSF \textsuperscript{$\ddagger$} \cite{koohpayegani2021mean} & \cite{koohpayegani2021mean} & 200 & \textbf{71.2} & \textbf{92.6} & \textbf{76.3} & \textbf{59.2} & \textbf{55.6}  & 73.2 & 88.7 & \textbf{92.7} & 92.0 & \textbf{77.9} \\
    ISD & - & 200 & 68.6 & 90.8 & 72.0 & 55.8 & 45.8 & 68.6 & \textbf{89.1} & 90.3 & 87.4 & 74.3 \\\hline
    \multicolumn{14}{|c|}{ResNet-18} \\\hline
    BYOL-asym \cite{grill2020bootstrap} & - & 200 & 55.0 & \textbf{83.4} & 59.3 & 48.2 & 26.6 & 65.4 & 74.1 & 82.7 & 82.3 & 64.1 \\
    MoCo v2 \cite{chen2020mocov2} & - & 200 & 56.7 & 83.0 & 59.7 & 48.8 & 30.4 & 64.4 & 70.1 & 80.5 & 83.1 & 64.1 \\
    ISD & - & 200 & \textbf{58.3} & 83.3 & \textbf{62.7} & \textbf{49.6} & \textbf{36.1} & \textbf{65.6} & \textbf{76.4} & \textbf{84.5} & \textbf{87.4} & \textbf{67.1} \\
    \hline
    \end{tabular}
    }
    \end{center}
    \caption{\textbf{Linear transfer evaluation: } We linear classifiers on top of frozen features for various downstream datasets. Hyperparameters are tuned individually for each method and the results are reported on the hold-out test sets. Our ResNet-18 is significantly better than other state-of-the-art SSL methods. ``rep.'' refers to the reproduction with our framework for a fair comparison. $\ddagger$: our concurrent work.}
    \label{tab:transfer_linear}
\end{table*}

\begin{table}[th!]
    \begin{center}
    \scalebox{0.93}{
    \begin{tabular}{|lccccc|}
        \hline
        \multirow{2}{*}{Method} & \multirow{2}{*}{Epochs} & \multicolumn{2}{c}{Top-1} & \multicolumn{2}{c|}{Top-5} \\
        & & 1\% & 10\% & 1\% & 10\% \\
        \hline
        \multicolumn{6}{|l|}{\textit{Entire network is fine-tuned.}} \\
        Supervised & & 25.4 & 56.4 & 48.4 & 80.4 \\
        % InstDisc \cite{wu2018unsupervised} & & - & - & 39.2 & 77.4\\
        PIRL \cite{misra2019self} & 800 & - & - & 57.2 & 83.8  \\
        CO2 \cite{wei2020co2} & 200 & - & - & 71.0 & 85.7 \\
        SimCLR \cite{chen2020simple} & 1000 & 48.3 & 65.6 & 75.5 & 87.8 \\
        InvP \cite{wang2020invp} & 800 & - & - & 78.2 & 88.7 \\
        BYOL \cite{grill2020bootstrap} & 1000 & 53.2 & 68.8 & 78.4 & 89.0 \\
        $\text{SwAV}^{\dagger}$ \cite{caron2020unsupervised} & 800 & \textbf{53.9} & \textbf{70.2} & \textbf{78.5} & \textbf{89.9} \\
        \hline
        \multicolumn{6}{|l|}{\textit{Only the linear layer is trained.}} \\
        $\text{BYOL}^{\ddag}$ \cite{grill2020bootstrap} & 1000 & 55.7 & \textbf{68.6} & 80.0 & \textbf{88.6} \\
        CompRess* \cite{abbasi2020compress} & 1K+130 & \textbf{59.7} & 67.0 & \textbf{82.3} & 87.5 \\[0.1cm]
        \hdashline
         & & & & & \\[-0.3cm]
        MoCo v2 \cite{chen2020mocov2} & 200 & 43.6 & 58.4 & 71.2 & 82.9 \\
        BYOL-asym \cite{grill2020bootstrap} & 200 & 47.9 & 61.3 & 74.6 & 84.7 \\
        ISD & 200 & \textbf{53.4} & \textbf{63.0} & \textbf{78.8} & \textbf{85.9} \\
        \hline
    \end{tabular}
    }
    \end{center}
    \caption{\textbf{Evaluation on limited labels ImageNet for ResNet-50: } We evaluate our model for the 1\% and 10\% ImageNet linear evaluation. Unlike other methods, we only train a single linear layer on top of the frozen backbone. We observe that our method is better than other state-of-the-art methods given similar computational budgets. * is compressed from ResNet-50x4}
    \label{tab:lim_sup}
\end{table}

{\bf ResNet-18 experiments:} 
Following CompRess \cite{abbasi2020compress}, we train our self-supervised ResNet-18 model with the initial learning rate set to $0.01$ and multiplied by $0.2$ at epochs 140 and 180. We follow \cite{grill2020bootstrap} and add a prediction layer for the student. The hidden and output dimensions of the prediction MLP layer are set to $512$. We do not have any projection layer for ResNet18. We use the same set of augmentations used in \cite{chen2020mocov2,chen2020simple,grill2020bootstrap}. We use the same temperature for teacher and student $\tau_s = \tau_t = 0.02$. The memory bank size is $128$K and momentum $m$ for teacher encoder is $0.999$. We choose these parameters based on ablations which can be found in the appendix. The results are shown in Tables \ref{tab:imagenet} and \ref{tab:transfer_linear}. Our model outperforms both baselines on full ImageNet linear and transfer linear benchmarks.

It is important to note that our method can be seen as a soft version of MoCo. So, ISD outperforming MoCo, empirically supports our main motivation of improving representations by smoothing the contrastive learning: not considering all negatives equally negative. %\ajinkya{add how we ran moco-v2 r18 baseline}

\begin{figure}[h!]
\begin{center}
\end{center}
  \includegraphics[width=1.0\linewidth]{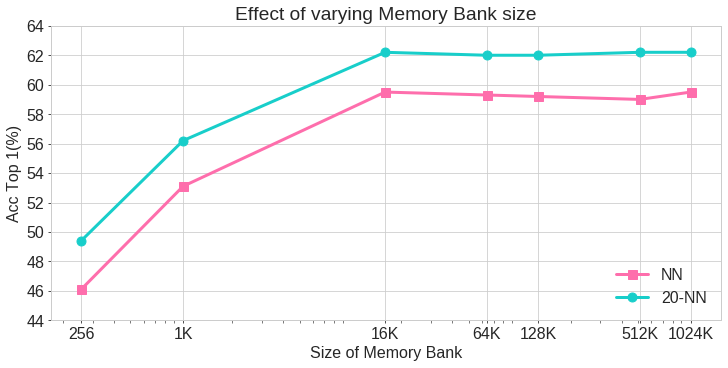}
  \caption{\textbf{Effect of Memory Bank Size:} We study the effect of memory bank size by varying from $256$ to $1024$K for ISD on ResNet-50 model.}
\label{fig:abl_isd_membank_size}
\end{figure}

{\bf ResNet-50 experiments:}
We train ResNet-50 with different settings than ResNet-18. We use same architecture and settings as ResNet-50 BYOL-asym. We use cos learning rate scheduler with initial learning rate of $0.05$. We use temperature of $\tau_s = 0.1$ for the student and $\tau_t = 0.01$ for the teacher. Memory bank size is $128$K, and momentum $m$ for teacher encoder is $0.99$. We study the effect of memory bank size in Figure \ref{fig:abl_isd_membank_size} which shows that memory bank size of even $16$K is on-par with $128$K and above. Additionally, inspired by \cite{sohn2020fixmatch}, we train our model with two different augmentation sets which we call it weak (random cropping and random horizontal flipping) and strong (same as \cite{chen2020mocov2,chen2020simple,grill2020bootstrap}). The teacher view uses weak augmentation while the student view uses the strong augmentation. We evaluate the effect of different augmentation in Table \ref{tab:abl_isd_aug}. ResNet50 results are shown in Tables \ref{tab:imagenet} and \ref{tab:transfer_linear}.

\begin{figure*}
\begin{center}
   \includegraphics[width=0.95\linewidth]{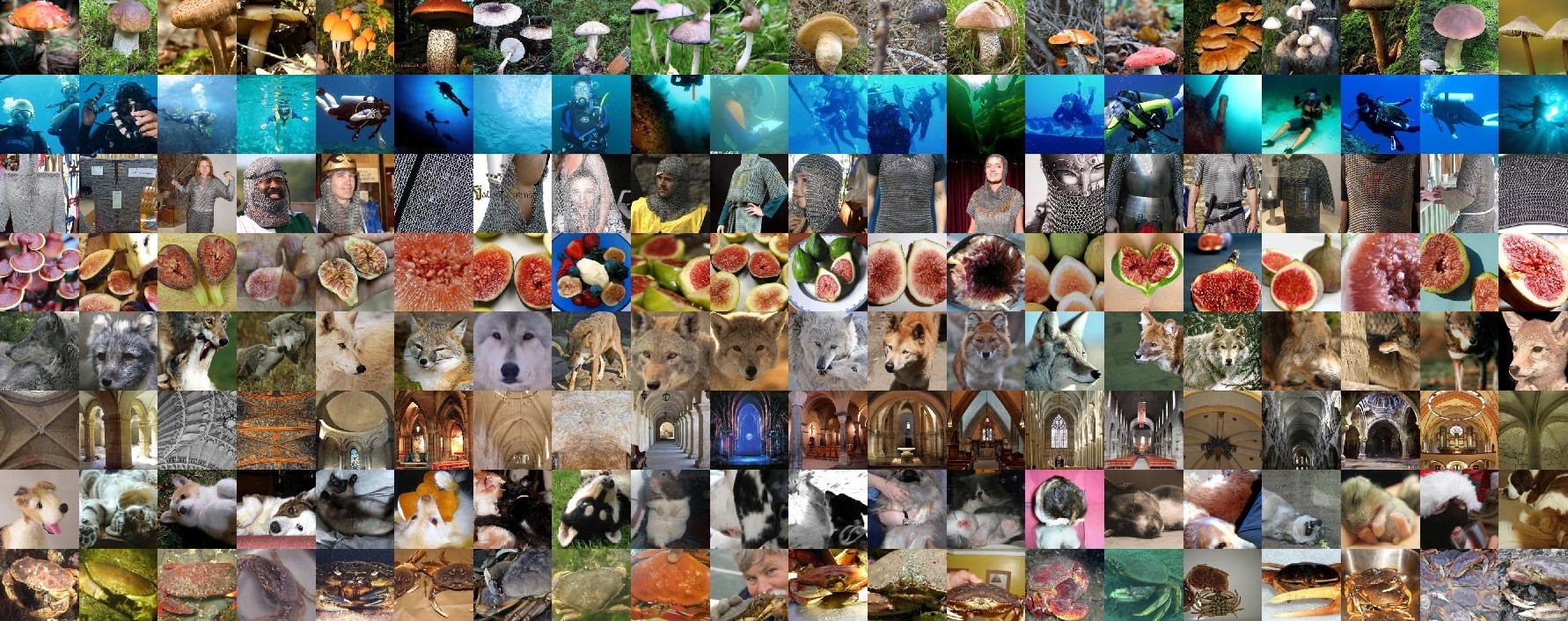}
   \caption{\textbf{Random Clusters:} We cluster ImageNet dataset into 1000 clusters using k-means and show random samples from random clusters. We did not do cherry-picking for this visualization. Each row corresponds to a cluster. Note that semantically similar images are clustered together. More results can be found in the appendix.
   }
   \vspace{-.2in}
\label{fig:clusters}
\end{center}
\end{figure*}

Tables \ref{tab:imagenet} and \ref{tab:transfer_linear} show the results on ImageNet and transfer learning settings respectively. Our method is comparable to SOTA SSL methods including BYOL in Linear and nearest neighbor evaluation on ImageNet. As mentioned earlier, we believe our method is more relaxed compared to BYOL as our method lets the embeddings of augmented images move as long as their similarity relationship with neighbors has not changed. 
% Hence, fine-tuning BYOL model using this more relaxed method (ISD) may result in a model that can generalize better to other tasks. Moreover, 
Table \ref{tab:lim_sup} shows our results when only limited labels are available in ImageNet dataset. Figure \ref{fig:clusters} shows random image samples from random clusters where each row corresponds to a cluster. Note that each row contains almost semantically similar images.

% Note that we have used the pretrained model for DCv2 and SwAV baselines provided by the authors, so they are using different set of augmentations.

\begin{figure*}
\begin{center}
\end{center}
   \includegraphics[width=1.0\linewidth]{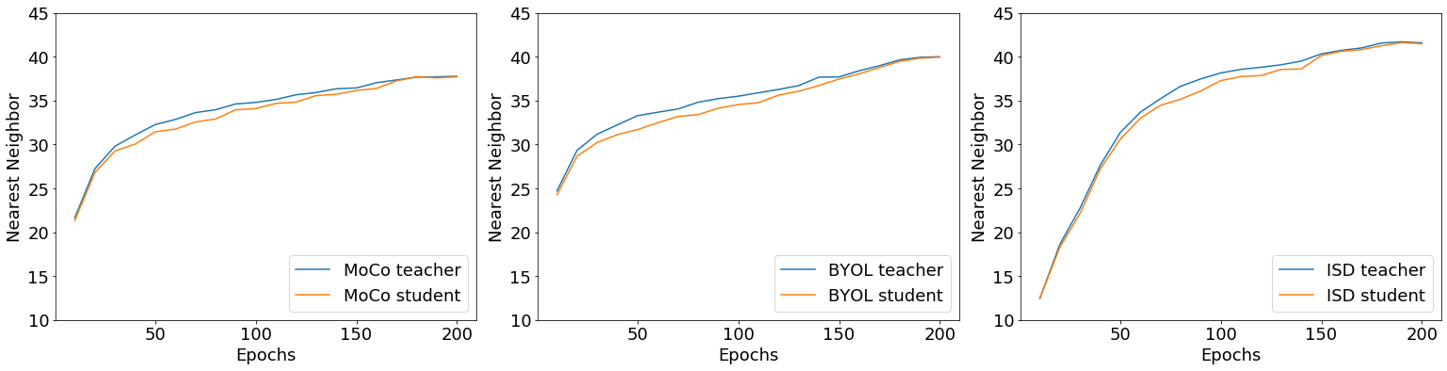}
   \caption{{\bf Evolution of teacher and student models:} Comparing the teacher and student ResNet18 models using Nearest Neighbor while training for MoCo, BYOL, and ISD methods. Interestingly, the teacher performs better than the student before shrinking the learning rate. Most previous works use the student as the final model which seems to be sub-optimal. We believe this is due to ensembling effect similar to \cite{tarvainen2017mean} and needs more investigation.}
   %\vspace{-.1in}
\label{fig:teacher_student}
\end{figure*}

{\bf Evolution of teacher and student models:}
In Figure \ref{fig:teacher_student}, for every 10 epoch of ResNet-18, we evaluate both teacher and student models for BYOL, MoCo, and ISD methods using nearest neighbor. For all methods, the teacher performs usually better than the student in the initial epochs when the learning rate is relatively large and then is very close to the student when it shrinks. This is interesting as we have not seen previous papers comparing the teacher with the student. This might happen since the teacher is a running average of the student so can be seen as an ensemble over many student networks similar to \cite{tarvainen2017mean}. We believe this deserves more investigation as future work.

{\bf Ablation study:} 
We varied the temperature for our method on ResNet-18 with 130 epochs and reported the results in Table \ref{tab:temp_abl}. Here, $LR=0.01$ and it is multiplied by $0.2$ at 90 and 120 epochs. Also, for more fair comparison with BYOL on ResNet18, we varied the learning rate and chose the best one for BYOL. Table \ref{tab:abl_byol_lr} shows the results of this experiment.

\begin{table*}
\begin{center}

\begin{tabular}{|l|c|c|c|c|c|c|c|c|c|c||c|}
\hline
Method & $D_1$& $D_2$& $D_3$& $D_4$& $D_5$ & $D_6$& $D_7$& $D_8$& $D_9$& $D_{10}$ & Mean\\
\hline
 \multicolumn{12}{|c|}{Evaluation On All 38 Categories}\\
\hline 
MoCo & 48.9 & 57.8 & 59.4  & 56.7  &  62.0 & 54.6 & 57.8 & 57.2 & 62.4 & 54.4 & 57.12\\
ISD & 49.6  & 57.9 & 61.9 & 58.6 & 62.3 & 56.3 & 57.8 & 58.1 & 62.5 & 55.3 & 58.03\\
\hline
Diff & +0.7  & +0.1 & +2.5 & +1.9 & +0.3&+1.7&0&+0.9&+0.1&+0.9 & +0.91 \\
\hline
\multicolumn{12}{|c|}{Evaluation Only on 30 Rare Categories}\\
\hline
MoCo & 44.7 & 52.3 & 57.3  & 53.1  &  57.7 & 50.7 & 51.1 & 51.9 & 58.9 & 59.8 & 53.75\\
ISD & 46.5  & 53.9 & 60.8 & 56.8 & 60.5 & 54.5 & 53.1 & 55.0 & 60.7 & 61.5 & 56.33\\
\hline
Diff & +1.7  & +1.6 & +3.5 & +3.7 &  +2.8 & +3.8 & +2.0 & +3.1 & +1.8 & +1.7 & +2.57\\
\hline
\end{tabular}
\end{center}
\caption{\textbf{Unbalanced dataset:} Nearest Neighbor (NN) results with ResNet-18 model for the unbalanced data when we consider all 38 categories and 30 small categories separately. We repeat the experiment 10 times with different random sets of 38 categories. NN is done on the validation set of ImageNet (which has uniform distribution) by searching the nearest neighbors among all ImageNet training data of those 38 categories (so the training data of NN also has uniform distribution). Hence, the whole evaluation is on balanced data to make sure we observe the effect of the unbalanced, ``unlabeled'' data only. ``Diff'' shows the improvement of our method over MoCo. Interestingly the improvement is bigger in the rare categories. This is aligned with out hypothesis that our method can handle unbalanced, unlabeled data better since it does not consider all negative images equally negative.}
\label{unbalanced_all}
\end{table*}

\begin{table}[h!]
    \begin{center}
    \begin{tabular}{|c|cc|cc|}
        \hline
        Method & Student & Teacher & NN & 20-NN\\
        & Aug. & Aug. & & \\\hline
        %ISD & strong & strong & 57.1 & 60.4\\
        %ISD & weak & strong & \textbf{58.5} & \textbf{61.5} \\
        ISD & weak & weak & 40.4 & 43.5 \\
        ISD & strong & weak & 22.9 & 26.3 \\
        ISD & strong & strong & 58.0 & 61.2 \\
        ISD & weak & strong & \textbf{59.2} & \textbf{62.0} \\
        \hline
    \end{tabular}
    \end{center}
    \caption{\textbf{Effect of augmentation strategies}: Effect of using weak or strong augmentations for ResNet-50 trained with 200 epochs. %\ajinkya{add BYOL for comparison here. BYOL w/w is very bad compared to ISD w/w.}
    }
    \label{tab:abl_isd_aug}
\end{table}

\begin{table}[h!]
    \begin{center}
    \begin{tabular}{|c|cccccc|}
        \hline
        $\tau$ & 0.003 & 0.007 & 0.01 & 0.02 & 0.04 & 0.06 \\
        \hline
        NN & 37.2 & 37.8 & 37.7 & \textbf{39.7} & 35.3 & 32.5 \\
        \hline
    \end{tabular}
    \end{center}
    \caption{\textbf{Effect of temperature: } Effect of changing temperature for our method ISD on ResNet-18 model.}
    \label{tab:temp_abl}
\end{table}

\begin{table}[h!]
    \begin{center}
    \begin{tabular}{|c|cccc|}
        \hline
        LR & 0.01 & 0.05 & 0.10 & 0.20 \\\hline
        NN & 37.3 & \textbf{40.0} & 38.6 & 37.3 \\
        \hline
    \end{tabular}
    \end{center}
    \caption{\textbf{Effect of learning rates for BYOL:} Comparison of different learning rates for BYOL on ResNet-18 with 200 epochs and cosine learning rate scheduler.}
    \label{tab:abl_byol_lr}
\end{table}

\subsection{Self-Supervised Learning on Unbalanced Dataset}
Most recent self-supervised learning methods are benchmarked by training on unlabeled ImageNet. However, we know that ImageNet has a particular bias of having an almost uniform distribution over the number of samples per category. We believe this bias does not exist in many real-world applications where the data is unbalanced: a few categories have a large number of samples while the rest of the data have a small number of samples. For instance, in self-driving car applications, it is really important to learn features for understanding rare scenes while most of the data is captured from repetitive safe highway scenes. Hence, we believe it is important to design and evaluate self-supervised learning methods for such unbalanced data. 

As mentioned earlier, since standard contrastive learning methods e.g., MoCo, consider all negative examples equally negative, when the query image is from a large category, it is possible to have multiple samples from the same category in the memory bank. Then, the contrastive loss in MoCo pushes their embeddings to be far apart as negative pairs. However, our method can handle such cases since our teacher assigns a soft label to the negative samples, so if an anchor example is very similar to the query, it will have high similarity and the student is optimized to reproduce such similarity.

To study our method on unbalanced data, we design a controlled setting to introduce the unbalanced data in the SSL training only and factor out its effect in the feature evaluation step. Hence, we sub-sample ImageNet data with 38 random categories where 8 categories are large (use all of almost 1300 images per category) and 30 categories are small (use only 100 images per category.) We train our SSL method and then evaluate by nearest neighbor (NN) classifier on the balanced validation data. To make sure that the feature evaluation is not affected by the unbalanced data, we keep both evaluation and the training data of NN search balanced, so for NN search, we use all ImageNet training images (almost $1300\times38$ images) for those 38 categories. 

We repeat the sampling of 38 categories 10 times to come up with 10 datasets and report the results for our method and also MoCo in Table \ref{unbalanced_all}. To measure the effect of the unbalanced data. we report the accuracy on all 38 categories and also on those 30 small categories only separately. Our method performs consistently better than MoCo, but more interestingly, the gap the improvement is larger when we evaluate on the 30 small categories only. We believe this empirically proves our hypothesis that our method may be able to handle unbalanced data more effectively. For a fair comparison, we train both our model and MoCo for 400 epochs with a memory bank size of $8192$ and cosine learning schedule.

\section{Related Work}

\textbf{Self-supervised learning:} The task of learning representations by solving a pretext task without using any supervised annotations is called self-supervised learning. Various pretext tasks like solving jigsaw puzzles \cite{noroozi2016unsupervised}, predicting rotations \cite{gidaris2018unsupervised}, counting the visual primitives \cite{noroozi2017representation}, filling up a missing patch \cite{pathak2016context}, predicting missing channels of input \cite{zhang2016colorful,zhang2017split}, and contrastive learning \cite{hadsell2006dimensionality,he2020momentum} are explored in the literature. We are proposing a novel pretext task based on iterative similarity based distillation.

{\bf Contrastive learning:} The task of learning unsupervised representations by contrasting the representations of an image with other images is called contrastive learning \cite{hadsell2006dimensionality}. Contrastive learning is essentially positive/negative classification where the positive and negative pairs of embeddings can be defined in various ways. In \cite{wu2018unsupervised,chen2020simple,he2020momentum}, the positive pairs are augmented views of the same image while the negative pairs are those of different images. In \cite{henaff2019data}, the positive pairs are a patch and context embeddings from the same image while the negatives pairs are a patch and context embedding from different images. In \cite{hjelm2018learning,bachman2019learning}, the positives pairs are global and local features from the same image while the negative pairs are global and local features from different images. In \cite{asano2020self,caron2020unsupervised,caron2018deep}, the positive pairs are members of the same cluster while the negative pairs are member of different clusters. We are different from these contrastive methods in that we do not consider all negatives equally: we calculate a soft labeling for the negatives using similarity of the data points. Also, we do not consider positive pairs directly: the comparison for the positive pairs is done through the similarity distillation. A few methods have attempted to fix false negative problem of contrastive learning by debiasing the loss \cite{chuang2020debiased} and by sampling the local neighborhood as positives \cite{wang2020invp}. We're different as we simply make the contrastive learning soft. \cite{huynh2020boosting} is a concurrent work that identifies some wrong negatives and cancels them in constrictive leaning.

\textbf{Knowledge Distillation:} The task of transferring the knowledge from one model to the other is called knowledge distillation \cite{hinton2015distilling,ba2014deep}. The knowledge from the teacher can be extracted and transferred in various ways. The knowledge in the activations of intermediate layers can be transferred through regression \cite{romero2015fitnet,yim2017gift,Zagoruyko2017AT}. While in most works the teacher is a deeper model and the student is a shallower model, in \cite{bagherinezhad2018label,furlanello2015born} both teacher and the student use the same architecture. Techniques from knowledge distillation can also be used in an unsupervised way to improve self-supervised learning \cite{abbasi2020compress,noroozi2018boosting,yan2020cluster}. Instead of using knowledge distillation for model compression or reducing the generalization gap, we use it iteratively to evolve the teacher and student together to learn rich representations from scratch. 

\textbf{Similarity based knowledge distillation:} While the above methods \cite{hinton2015distilling,ba2014deep,romero2015fitnet,yim2017gift,Zagoruyko2017AT} only extract the information about a single data point from the teacher, similarity based distillation methods \cite{passalis2018learning,peng2019correlation,park2019relational,tung2019similarity,ahn2019variational,Tian2020Contrastive,abbasi2020compress,fang2021seed} represent the knowledge from the teacher in terms of similarities between data points. CompRess \cite{abbasi2020compress} and SEED \cite{fang2021seed} are the closest related work to our method that uses similarity based distillation to compresses a large self-supervised model to a smaller one. Our method is different as we use similarity based distillation to iteratively distill an evolving teacher to a student.

\textbf{Consistency regularization:} Consistency regularization is a method of regularization that seeks to make the output of a model consistent across small perturbations in either the input \cite{miyato2018virtual} or the model parameters \cite{tarvainen2017mean}. Recently, BYOL \cite{grill2020bootstrap} applied a variant of Mean Teachers \cite{tarvainen2017mean} for self-supervised learning. %Note that BYOL \cite{grill2020bootstrap} proposed an architecture that prevents the collapse of the model despite not having any contrastive or supervised loss component. 
Our method uses a variation of this idea through similarity based loss rather than regression loss defined on data points individually. %\soroush{Additionally, \cite{wei2020co2} use similar loss as \cite{abbasi2020compress} as a regularize for training MoCo. Unlike \cite{wei2020co2}, we don't train our model with any contrastive loss and we are able to train such a model from scratch.}
\cite{wei2020co2} is probably the closet to ours that uses a loss similar to our as a regularizer in addition to the MoCo loss. Our method is different as we optimize our loss from scratch as the  main objective without adding it to another method. Also, our method benefits from different temperatures for the teacher and student networks which is inspired by \cite{fang2021seed}.

\section{Conclusion}
We introduce ISD, a novel self-supervised learning method. It is a variation of contrastive learning (e.g., MoCo) in which negative samples are not all treated equally. The similarity between images in the teacher's embedding space determines how much each anchor image should be contrasted with. Our extensive experiments show that our method performs comparable to the state-of-the-art SSL methods on ImageNet, transfer learning tasks, and when the unlabeled data is unbalanced. %\ajinkya{edit after final results. has sota claims.} \\

\noindent {\bf Acknowledgment:} 
%\soroush{acknowledge AC for the Accept!}
This material is based upon work partially supported by the United States Air Force under Contract No. FA8750‐19‐C‐0098, funding from SAP SE, and also NSF grant numbers 1845216 and 1920079. Any opinions, findings, and conclusions or recommendations expressed in this material are those of the authors and do not necessarily reflect the views of the United States Air Force, DARPA, or other funding agencies. 

%-------------------------------------------------------------------------

{\small
\bibliographystyle{ieee_fullname}
\bibliography{egbib}

\begin{thebibliography}{10}\itemsep=-1pt

\bibitem{abbasi2020compress}
Soroush Abbasi~Koohpayegani, Ajinkya Tejankar, and Hamed Pirsiavash.
\newblock Compress: Self-supervised learning by compressing representations.
\newblock {\em Advances in Neural Information Processing Systems}, 33, 2020.

\bibitem{ahn2019variational}
Sungsoo Ahn, Shell~Xu Hu, Andreas Damianou, Neil~D Lawrence, and Zhenwen Dai.
\newblock Variational information distillation for knowledge transfer.
\newblock In {\em Proceedings of the IEEE Conference on Computer Vision and
  Pattern Recognition}, pages 9163--9171, 2019.

\bibitem{ba2014deep}
Jimmy Ba and Rich Caruana.
\newblock Do deep nets really need to be deep?
\newblock In {\em Advances in neural information processing systems}, pages
  2654--2662, 2014.

\bibitem{bachman2019learning}
Philip Bachman, R~Devon Hjelm, and William Buchwalter.
\newblock Learning representations by maximizing mutual information across
  views.
\newblock In {\em Advances in Neural Information Processing Systems}, pages
  15535--15545, 2019.

\bibitem{bagherinezhad2018label}
Hessam Bagherinezhad, Maxwell Horton, Mohammad Rastegari, and Ali Farhadi.
\newblock Label refinery: Improving imagenet classification through label
  progression.
\newblock {\em arXiv preprint arXiv:1805.02641}, 2018.

\bibitem{food101}
Lukas Bossard, Matthieu Guillaumin, and Luc Van~Gool.
\newblock Food-101 -- mining discriminative components with random forests.
\newblock In {\em European Conference on Computer Vision}, 2014.

\bibitem{caron2018deep}
Mathilde Caron, Piotr Bojanowski, Armand Joulin, and Matthijs Douze.
\newblock Deep clustering for unsupervised learning of visual features.
\newblock In {\em Proceedings of the European Conference on Computer Vision
  (ECCV)}, pages 132--149, 2018.

\bibitem{caron2020unsupervised}
Mathilde Caron, Ishan Misra, Julien Mairal, Priya Goyal, Piotr Bojanowski, and
  Armand Joulin.
\newblock Unsupervised learning of visual features by contrasting cluster
  assignments.
\newblock {\em arXiv preprint arXiv:2006.09882}, 2020.

\bibitem{chen2020simple}
Ting Chen, Simon Kornblith, Mohammad Norouzi, and Geoffrey Hinton.
\newblock A simple framework for contrastive learning of visual
  representations.
\newblock {\em arXiv preprint arXiv:2002.05709}, 2020.

\bibitem{chen2020mocov2}
Xinlei Chen, Haoqi Fan, Ross Girshick, and Kaiming He.
\newblock Improved baselines with momentum contrastive learning.
\newblock {\em arXiv preprint arXiv:2003.04297}, 2020.

\bibitem{chen2020exploring}
Xinlei Chen and Kaiming He.
\newblock Exploring simple siamese representation learning, 2020.

\bibitem{chuang2020debiased}
Ching-Yao Chuang, Joshua Robinson, Yen-Chen Lin, Antonio Torralba, and Stefanie
  Jegelka.
\newblock Debiased contrastive learning.
\newblock In {\em Advances in Neural Information Processing Systems}, 2020.

\bibitem{dtd}
Mircea Cimpoi, Subhransu Maji, Iasonas Kokkinos, Sammy Mohamed, and Andrea
  Vedaldi.
\newblock Describing textures in the wild.
\newblock In {\em Computer Vision and Pattern Recognition}, 2014.

\bibitem{fang2021seed}
Zhiyuan Fang, Jianfeng Wang, Lijuan Wang, Lei Zhang, Yezhou Yang, and Zicheng
  Liu.
\newblock Seed: Self-supervised distillation for visual representation.
\newblock In {\em International Conference on Learning Representations}, 2021.

\bibitem{caltech101}
Li Fei-Fei, Rob Fergus, and Pietro Perona.
\newblock Learning generative visual models from few training examples: An
  incremental bayesian approach tested on 101 object categories.
\newblock {\em Computer Vision and Pattern Recognition Workshop}, 2004.

\bibitem{furlanello2015born}
Tommaso Furlanello, Zachary~Chase Lipton, Michael Tschannen, Laurent Itti, and
  Anima Anandkumar.
\newblock Born-again neural networks.
\newblock In Jennifer~G. Dy and Andreas Krause, editors, {\em Proceedings of
  the 35th International Conference on Machine Learning, {ICML} 2018,
  Stockholmsm{\"{a}}ssan, Stockholm, Sweden, July 10-15, 2018}, volume~80 of
  {\em Proceedings of Machine Learning Research}, pages 1602--1611. {PMLR},
  2018.

\bibitem{gidaris2018unsupervised}
Spyros Gidaris, Praveer Singh, and Nikos Komodakis.
\newblock Unsupervised representation learning by predicting image rotations.
\newblock In {\em International Conference on Learning Representations}, 2018.

\bibitem{grill2020bootstrap}
Jean-Bastien Grill, Florian Strub, Florent Altch{\'e}, Corentin Tallec,
  Pierre~H Richemond, Elena Buchatskaya, Carl Doersch, Bernardo~Avila Pires,
  Zhaohan~Daniel Guo, Mohammad~Gheshlaghi Azar, et~al.
\newblock Bootstrap your own latent: A new approach to self-supervised
  learning.
\newblock {\em arXiv preprint arXiv:2006.07733}, 2020.

\bibitem{hadsell2006dimensionality}
Raia Hadsell, Sumit Chopra, and Yann LeCun.
\newblock Dimensionality reduction by learning an invariant mapping.
\newblock In {\em 2006 IEEE Computer Society Conference on Computer Vision and
  Pattern Recognition (CVPR'06)}, volume~2, pages 1735--1742. IEEE, 2006.

\bibitem{he2020momentum}
Kaiming He, Haoqi Fan, Yuxin Wu, Saining Xie, and Ross Girshick.
\newblock Momentum contrast for unsupervised visual representation learning.
\newblock In {\em Proceedings of the IEEE/CVF Conference on Computer Vision and
  Pattern Recognition}, pages 9729--9738, 2020.

\bibitem{henaff2019data}
Olivier~J H{\'e}naff, Aravind Srinivas, Jeffrey De~Fauw, Ali Razavi, Carl
  Doersch, SM Eslami, and Aaron van~den Oord.
\newblock Data-efficient image recognition with contrastive predictive coding.
\newblock {\em arXiv preprint arXiv:1905.09272}, 2019.

\bibitem{hinton2015distilling}
Geoffrey Hinton, Oriol Vinyals, and Jeff Dean.
\newblock Distilling the knowledge in a neural network.
\newblock {\em arXiv preprint arXiv:1503.02531}, 2015.

\bibitem{hjelm2018learning}
R~Devon Hjelm, Alex Fedorov, Samuel Lavoie-Marchildon, Karan Grewal, Phil
  Bachman, Adam Trischler, and Yoshua Bengio.
\newblock Learning deep representations by mutual information estimation and
  maximization.
\newblock In {\em International Conference on Learning Representations}, 2019.

\bibitem{huynh2020boosting}
Tri Huynh, Simon Kornblith, Matthew~R. Walter, Michael Maire, and Maryam
  Khademi.
\newblock Boosting contrastive self-supervised learning with false negative
  cancellation, 2020.

\bibitem{koohpayegani2021mean}
Soroush~Abbasi Koohpayegani, Ajinkya Tejankar, and Hamed Pirsiavash.
\newblock Mean shift for self-supervised learning, 2021.

\bibitem{carsdataset}
Jonathan Krause, Michael Stark, Jia Deng, and Li Fei-Fei.
\newblock {3D} object representations for fine-grained categorization.
\newblock In {\em Workshop on 3D Representation and Recognition}, Sydney,
  Australia, 2013.

\bibitem{cifar}
Alex Krizhevsky.
\newblock Learning multiple layers of features from tiny images.
\newblock Technical report, University of Toronto, 2009.

\bibitem{misra2019self}
Ishan Misra and Laurens van~der Maaten.
\newblock Self-supervised learning of pretext-invariant representations.
\newblock {\em arXiv preprint arXiv:1912.01991}, 2019.

\bibitem{miyato2018virtual}
Takeru Miyato, Shin-ichi Maeda, Masanori Koyama, and Shin Ishii.
\newblock Virtual adversarial training: a regularization method for supervised
  and semi-supervised learning.
\newblock {\em IEEE transactions on pattern analysis and machine intelligence},
  41(8):1979--1993, 2018.

\bibitem{flowers}
Maria-Elena Nilsback and Andrew Zisserman.
\newblock Automated flower classification over a large number of classes.
\newblock In {\em Indian Conference on Computer Vision, Graphics and Image
  Processing}, 2008.

\bibitem{noroozi2016unsupervised}
Mehdi Noroozi and Paolo Favaro.
\newblock Unsupervised learning of visual representations by solving jigsaw
  puzzles.
\newblock In {\em European Conference on Computer Vision}, pages 69--84.
  Springer, 2016.

\bibitem{noroozi2017representation}
Mehdi Noroozi, Hamed Pirsiavash, and Paolo Favaro.
\newblock Representation learning by learning to count.
\newblock In {\em Proceedings of the IEEE International Conference on Computer
  Vision}, pages 5898--5906, 2017.

\bibitem{noroozi2018boosting}
Mehdi Noroozi, Ananth Vinjimoor, Paolo Favaro, and Hamed Pirsiavash.
\newblock Boosting self-supervised learning via knowledge transfer.
\newblock In {\em Proceedings of the IEEE Conference on Computer Vision and
  Pattern Recognition}, pages 9359--9367, 2018.

\bibitem{park2019relational}
Wonpyo Park, Dongju Kim, Yan Lu, and Minsu Cho.
\newblock Relational knowledge distillation.
\newblock In {\em Proceedings of the IEEE Conference on Computer Vision and
  Pattern Recognition}, pages 3967--3976, 2019.

\bibitem{pets}
O.~M. Parkhi, A. Vedaldi, A. Zisserman, and C.~V. Jawahar.
\newblock Cats and dogs.
\newblock In {\em Computer Vision and Pattern Recognition}, 2012.

\bibitem{passalis2018learning}
Nikolaos Passalis and Anastasios Tefas.
\newblock Learning deep representations with probabilistic knowledge transfer.
\newblock In {\em Proceedings of the European Conference on Computer Vision
  (ECCV)}, pages 268--284, 2018.

\bibitem{pathak2016context}
Deepak Pathak, Philipp Krahenbuhl, Jeff Donahue, Trevor Darrell, and Alexei~A
  Efros.
\newblock Context encoders: Feature learning by inpainting.
\newblock In {\em Proceedings of the IEEE conference on computer vision and
  pattern recognition}, pages 2536--2544, 2016.

\bibitem{peng2019correlation}
Baoyun Peng, Xiao Jin, Jiaheng Liu, Dongsheng Li, Yichao Wu, Yu Liu, Shunfeng
  Zhou, and Zhaoning Zhang.
\newblock Correlation congruence for knowledge distillation.
\newblock In {\em Proceedings of the IEEE International Conference on Computer
  Vision}, pages 5007--5016, 2019.

\bibitem{romero2015fitnet}
Adriana Romero, Nicolas Ballas, Samira~Ebrahimi Kahou, Antoine Chassang, Carlo
  Gatta, and Yoshua Bengio.
\newblock Fitnets: Hints for thin deep nets.
\newblock In Yoshua Bengio and Yann LeCun, editors, {\em 3rd International
  Conference on Learning Representations, {ICLR} 2015, San Diego, CA, USA, May
  7-9, 2015, Conference Track Proceedings}, 2015.

\bibitem{sohn2020fixmatch}
Kihyuk Sohn, David Berthelot, Nicholas Carlini, Zizhao Zhang, Han Zhang,
  Colin~A Raffel, Ekin~Dogus Cubuk, Alexey Kurakin, and Chun-Liang Li.
\newblock Fixmatch: Simplifying semi-supervised learning with consistency and
  confidence.
\newblock {\em Advances in Neural Information Processing Systems}, 33, 2020.

\bibitem{tarvainen2017mean}
Antti Tarvainen and Harri Valpola.
\newblock Mean teachers are better role models: Weight-averaged consistency
  targets improve semi-supervised deep learning results.
\newblock In {\em Advances in neural information processing systems}, pages
  1195--1204, 2017.

\bibitem{Tian2020Contrastive}
Yonglong Tian, Dilip Krishnan, and Phillip Isola.
\newblock Contrastive representation distillation.
\newblock In {\em International Conference on Learning Representations}, 2020.

\bibitem{tung2019similarity}
Frederick Tung and Greg Mori.
\newblock Similarity-preserving knowledge distillation.
\newblock In {\em Proceedings of the IEEE International Conference on Computer
  Vision}, pages 1365--1374, 2019.

\bibitem{wang2020invp}
Feng Wang, Huaping Liu, Di Guo, and Sun Fuchun.
\newblock Unsupervised representation learning by invariance propagation.
\newblock In {\em Advances in Neural Information Processing Systems}, 2020.

\bibitem{wei2020co2}
Chen Wei, Huiyu Wang, Wei Shen, and Alan Yuille.
\newblock Co2: Consistent contrast for unsupervised visual representation
  learning.
\newblock {\em arXiv preprint arXiv:2010.02217}, 2020.

\bibitem{wu2018unsupervised}
Zhirong Wu, Yuanjun Xiong, Stella~X Yu, and Dahua Lin.
\newblock Unsupervised feature learning via non-parametric instance
  discrimination.
\newblock In {\em Proceedings of the IEEE Conference on Computer Vision and
  Pattern Recognition}, pages 3733--3742, 2018.

\bibitem{sun397}
Jianxiong {Xiao}, James {Hays}, Krista~A. {Ehinger}, Aude {Oliva}, and Antonio
  {Torralba}.
\newblock Sun database: Large-scale scene recognition from abbey to zoo.
\newblock In {\em Computer Vision and Pattern Recognition}, 2010.

\bibitem{yan2020cluster}
Xueting Yan, Ishan Misra, Abhinav Gupta, Deepti Ghadiyaram, and Dhruv Mahajan.
\newblock Clusterfit: Improving generalization of visual representations.
\newblock In {\em CVPR}, 2020.

\bibitem{yim2017gift}
Junho Yim, Donggyu Joo, Jihoon Bae, and Junmo Kim.
\newblock A gift from knowledge distillation: Fast optimization, network
  minimization and transfer learning.
\newblock In {\em Proceedings of the IEEE Conference on Computer Vision and
  Pattern Recognition}, pages 4133--4141, 2017.

\bibitem{asano2020self}
Asano YM., Rupprecht C., and Vedaldi A.
\newblock Self-labelling via simultaneous clustering and representation
  learning.
\newblock In {\em International Conference on Learning Representations}, 2020.

\bibitem{Zagoruyko2017AT}
Sergey Zagoruyko and Nikos Komodakis.
\newblock Paying more attention to attention: Improving the performance of
  convolutional neural networks via attention transfer.
\newblock In {\em ICLR}, 2017.

\bibitem{zhang2016colorful}
Richard Zhang, Phillip Isola, and Alexei~A Efros.
\newblock Colorful image colorization.
\newblock In {\em European conference on computer vision}, pages 649--666.
  Springer, 2016.

\bibitem{zhang2017split}
Richard Zhang, Phillip Isola, and Alexei~A Efros.
\newblock Split-brain autoencoders: Unsupervised learning by cross-channel
  prediction.
\newblock In {\em Proceedings of the IEEE Conference on Computer Vision and
  Pattern Recognition}, pages 1058--1067, 2017.

\end{thebibliography}
}

\clearpage
\appendix
\onecolumn

\setcounter{figure}{0}
\renewcommand\thefigure{A\arabic{figure}}
\setcounter{table}{0}
\renewcommand\thetable{A\arabic{table}}

\section*{Appendix}

\textbf{Transfer evaluation training details:} we freeze the backbone and forward train set images without augmentation (resize shorter side to 256, take a center crop of size 224, and normalize with  ImageNet  statistics). Then we train a linear layer on top of extracted features. We split each dataset to train, validation, and test set. We search for best lr in 10 log spaced values between -3 and 0 and weight decay in 9 log spaced values between -10 and -2, then we train linear layer with best parameters on train+validation set and evaluate it on test set.

\begin{table*}[h]
    \begin{center}
    \scalebox{0.95}{
    \begin{tabular}{lrrrrrrr}
        \toprule
        Dataset & Classes & Train samples & Val samples & Test samples & Accuracy measure & Test provided \\
        \midrule
        Food101 [6] & 101 & 68175 & 7575 & 25250 & Top-1 accuracy & - \\
        CIFAR-10 [25] & 10 & 49500 & 500 & 10000 & Top-1 accuracy & - \\
        CIFAR-100 [25] & 100 & 45000 & 5000 & 10000 & Top-1 accuracy & -\\
        Sun397 (split 1) [45] & 397 & 15880 & 3970 & 19850 & Top-1 accuracy & - \\
        Cars [24] & 196 & 6509 & 1635 & 8041 & Top-1 accuracy & - \\
        DTD (split 1) [13] & 47 & 1880 & 1880 & 1880 & Top-1 accuracy & Yes \\
        Pets [33] & 37 & 2940 & 740 & 3669 & Mean per-class accuracy & - \\
        Caltech-101 [15] & 101 & 2550 & 510 & 6084 & Mean per-class accuracy & - \\
        Flowers [28] & 102 & 1020 & 1020 & 6149 & Mean per-class accuracy & Yes \\
        \bottomrule
    \end{tabular}
    }
    \end{center}
    \caption{The train, val, and test split sizes for transfer datasets are listed above. \textbf{Test split: } For DTD and Flowers datasets, we use the provided test sets. Otherwise, in case of Sun397, Cars, CIFAR-10, CIFAR-100, Food101, and Pets datasets, the val set provided in the dataset is used as the hold-out test set. Also, for Caltech-101, the hold-out test set is created by randomly sampling 30 images/class from the train set. \textbf{Val split: } For DTD and Flowers datasets, we use the provided val sets. Otherwise, the val set is created by a randomly sampled subset of the train set is used as the val set. We report the strategy for splitting val sets for different datasets: 5 samples/class for Caltech-101, 20\% samples/class for Cars, 50 samples/class for CIFAR-100, 50 samples/class for CIFAR-10. 75 samples/class for Food101, 20 samples/class for Pets, 10 samples/class for Sun397. We attempt to be as close to the details provided in BYOL [18] as possible.}
    \label{tab:transfer_dset_details}
\end{table*}

\begin{table}
    \centering
    \scalebox{0.9}{
    \begin{tabular}{|lllcccccc|}
        \hline
        & LR & $m$ & Aug. & Proj. & $\tau_t$ & $\tau_s$ & NN & 20-NN \\
        % Sched. & & & & temp & temp & & \\
        \hline
        1 & step & 0.999 & s/s & \ding{55} & 0.02 & 0.02 & 41.5 & 46.6 \\
        2 & step & 0.999 & w/s & \ding{55} & 0.02 & 0.02 & 41.7 & 46.7 \\
        3 & step & 0.99 & s/s & \ding{55} & 0.02 & 0.02 & 40.2 & 45.2 \\
        4 & cosine & 0.999 & s/s & \ding{55} & 0.02 & 0.02 & 40.9 & 45.7 \\
        5 & cosine & 0.99 & w/s & \ding{51} & 0.02 & 0.02 & 34.6 & 38.3 \\
        6 & cosine & 0.99 & w/s & \ding{51} & 0.02 & 0.2 & 39.4 & 44.4 \\
        7 & cosine & 0.99 & w/s & \ding{51} & 0.01 & 0.1 & 31.7 & 37.3 \\
        8 & step & 0.999 & w/s & \ding{55} & 0.02 & 0.2 & 6.0 & 8.0 \\
        9 & step & 0.999 & w/s & \ding{55} & 0.02 & 0.5 & 5.5 & 7.0 \\
        10 & step & 0.999 & w/s & \ding{55} & 0.03 & 0.3 & 2.9 & 3.9 \\
        \hline
    \end{tabular}
    }
    \caption{We explore various hyper-parameters for ResNet-18 model. ``s/s'' refers to the setting where both views use strong augmentation while in ``w/s'' the teacher view uses weak augmentation while the student view uses strong augmentation. The projection layer is a 2-layer MLP with 1024 as hidden dim and 128 as the output dim. There is a BatchNorm layer followed by ReLU between the two layers. In step learning rate decay, the LR is reduced by a factor of 0.2 at 140 and 180 epochs. The training happens for 200 epochs. The 1st row uses the same setting as the ResNet-18 model in the main paper. The 6th row uses the same settings as the ResNet-50 model in the main paper.}
    \label{tab:my_label}
\end{table}

\begin{figure*}[!b]
\begin{center}
   \includegraphics[width=0.85\linewidth]{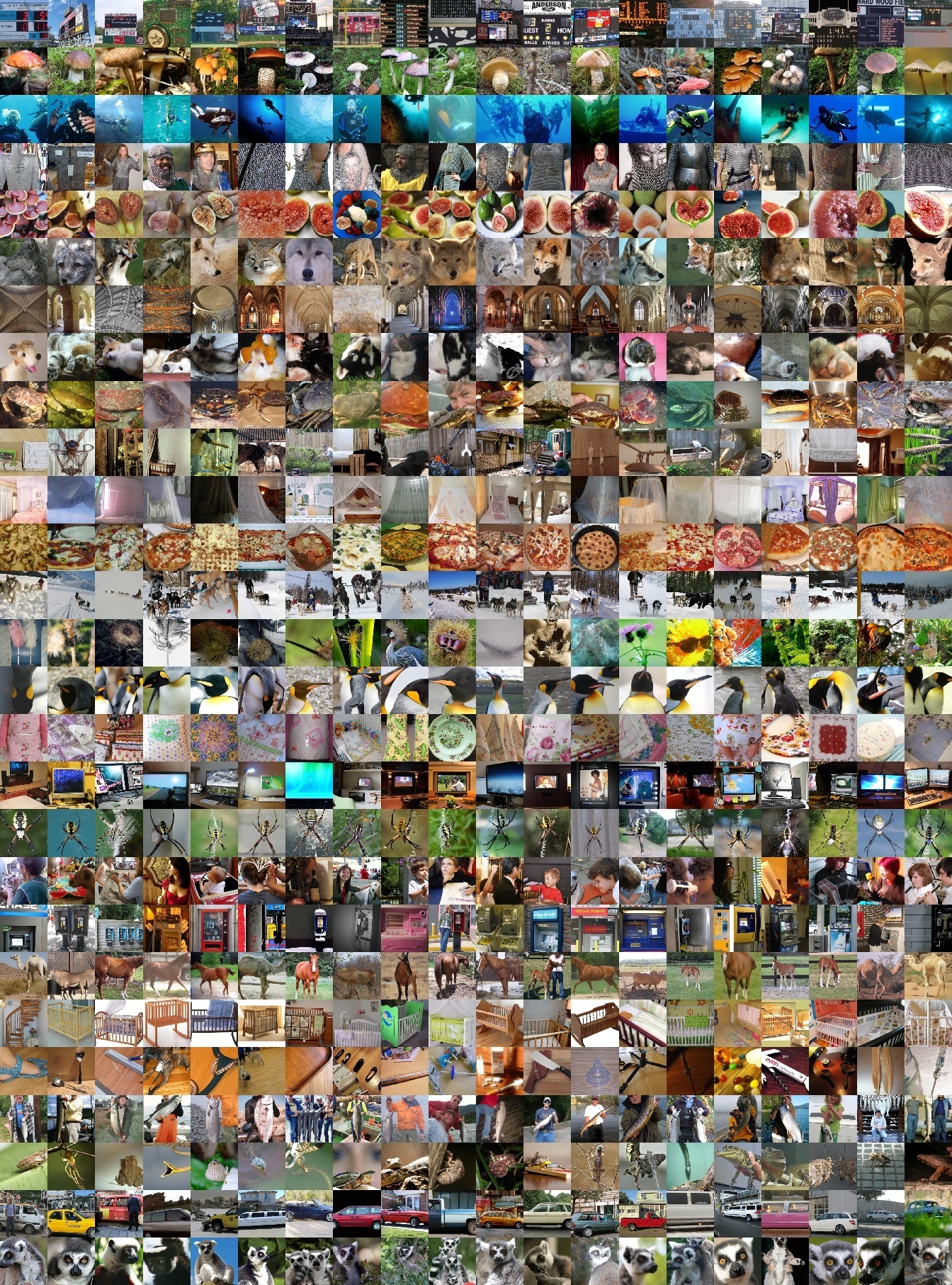}
   \caption{\textbf{Random Clusters:} We cluster ImageNet dataset into 1000 clusters using k-means and show random samples from random clusters. We have no cherry-picking for this visualization. Interestingly, images from each row(each cluster) are semantically similar. 
   }
   \vspace{-.2in}
\label{fig:appendix_clusters}
\end{center}
\end{figure*}

\end{document}